\documentclass[conference]{IEEEtran}
\IEEEoverridecommandlockouts
\usepackage{cite}
\usepackage{amsmath,amssymb,amsfonts}
\usepackage{algorithmic}
\usepackage{graphicx}
\usepackage{textcomp}
\usepackage[inline]{enumitem}
\usepackage{xcolor}
\usepackage{soul, color}
\usepackage{witharrows}
\usepackage[flushleft]{threeparttable}
\usepackage{tablefootnote}
\usepackage{array}
\newcolumntype{P}[1]{>{\centering\arraybackslash}p{#1}}
\usepackage{multicol}
\usepackage{multirow}
\usepackage{tabulary}
\usepackage{colortbl}
\usepackage{makecell}
\usepackage{booktabs}
\usepackage{subcaption}
\usepackage{stmaryrd}
\usepackage{float}
\usepackage{pifont}
\usepackage{arydshln}
\colorlet{mygray}{gray!15!white}

\usepackage{url,hyperref,microtype}
\hypersetup{
    colorlinks=true,
    citecolor=blue,
    linkcolor=blue,
    filecolor=magenta,      
    urlcolor=black,
}

\def\BibTeX{{\rm B\kern-.05em{\sc i\kern-.025em b}\kern-.08em
    T\kern-.1667em\lower.7ex\hbox{E}\kern-.125emX}}
\begin{document}

\title{A Lightweight Transformer for Pain Recognition from Brain Activity}

\author{\IEEEauthorblockN{Stefanos Gkikas}
\IEEEauthorblockA{\textit{Honda Research Institute Japan} \\
Wako City, Japan \\
stefanos.gkikas@jp.honda-ri.com}
\and
\IEEEauthorblockN{Christian Arzate Cruz}
\IEEEauthorblockA{\textit{Honda Research Institute Japan} \\
Wako City, Japan  \\
christian.arzate@jp.honda-ri.com}
\and
\IEEEauthorblockN{Yu Fang}
\IEEEauthorblockA{\textit{Honda Research Institute Japan} \\
Wako City, Japan \\
yu.fang@jp.honda-ri.com}
\and
\IEEEauthorblockN{Lu Cao}
\IEEEauthorblockA{\textit{Honda Research Institute Japan} \\
Wako City, Japan \\
lu.cao@jp.honda-ri.com}
\and
\IEEEauthorblockN{Muhammad Umar Khan}
\IEEEauthorblockA{\textit{BioSIS (Biosensing \& Intelligent Systems) Lab}\\ 
\textit{Centre for Intelligent Computing and Systems} \\
\textit{University of Canberra}\\
Canberra, Australia\\
umar.khan@canberra.edu.au}
\and
\IEEEauthorblockN{Thomas Kassiotis}
\IEEEauthorblockA{\textit{Department of Electronic Engineering} \\
\textit{Hellenic Mediterranean University}\\
Chania, Greece \\
ddk305@edu.hmu.gr}
\and
\IEEEauthorblockN{Giorgos Giannakakis}
\IEEEauthorblockA{\textit{Department of Electronic Engineering} \\
\textit{Hellenic Mediterranean University}\\
Chania, Greece \\
ggian@hmu.gr}
\and
\IEEEauthorblockN{Raul Fernandez Rojas}
\IEEEauthorblockA{\textit{BioSIS (Biosensing \& Intelligent Systems) Lab}\\ 
\textit{Centre for Intelligent Computing and Systems} \\
\textit{University of Canberra}\\
Canberra, Australia\\
raul.fernandezrojas@canberra.edu.au}
\and
\IEEEauthorblockN{Randy Gomez}
\IEEEauthorblockA{\textit{Honda Research Institute Japan} \\
Wako City, Japan \\
r.gomez@jp.honda-ri.com}
}

\maketitle

\begin{abstract}

Accurate and reliable pain assessment is an important clinical challenge, especially for patients who cannot communicate their pain directly. In this study, we present a lightweight transformer architecture for pain recognition that leverages multiple fNIRS signal representations and a unified tokenization framework. The proposed architecture enables the integration of heterogeneous inputs in a shared latent space, without requiring separate processing branches or additional architectural complexity. It further employs a structured latent segmentation strategy to capture both local aggregation and global interaction across spatial, temporal, and time--frequency features.
Evaluated on the \textit{AI4Pain} dataset, combining raw waveform and power spectral density fNIRS inputs, the proposed model achieves strong recognition performance across a wide range of computational configurations and, at the same time, maintains efficient inference on both GPU and CPU hardware.

\end{abstract}

\begin{IEEEkeywords}
Pain assessment, deep learning, neural signals, data fusion
\end{IEEEkeywords}

\section{Introduction}

Pain plays a fundamental protective role by alerting the body to possible tissue damage or disease, and supporting physiological integrity \cite{santiago_2022}. By nature, pain is a subjective and multidimensional experience that includes nociceptive, sensory, affective, and cognitive components \cite{marchand_2024}.
Pain has been characterized as a \textit{\textquotedblleft silent public health crisis \textquotedblright}\cite{katzman_gallagher_2024} because of the high prevalence and its impact on society.
Opioid analgesics remain the most frequently prescribed treatment for pain \cite{kaye_jones_2017}, despite their well-documented risks of addiction, overdose \cite{stampas_pedroza_2020}, and side effects such as sedation, depression, anxiety, and nausea that severely impair daily functioning and quality of life \cite{benyamin_trescot_2008}.
The subjective and multifaceted nature of pain complicates the accurate assessment in both clinical practice and research. The dependence on self-reported measurement can reduce treatment accuracy and also may contribute to the overprescription and misuse of analgesics \cite{meehan_mcrae_1995}.

\IEEEpubidadjcol

These challenges are especially pronounced in patients who cannot communicate effectively or directly \cite{puntilo_staannard_2022}. In critically ill adults, pain is often underrecognized and insufficiently treated, because the objective and standardized assessment tools are limited for this category of patients\cite{meehan_mcrae_1995}. Cancer-related pain is especially common, particularly in advanced stages of the disease \cite{snijders_brom_2022}.
In addition, pain directly affects attention, functional performance, and cognitive processing speed \cite{khera_rangasamy_2021}. For instance, Li \textit{et al.} \cite{li_lyu_2025} showed that pain disrupts cognitive control by impairing expectancy processing and decision formation. Moore \textit{et al.} \cite{moore_meints_2019} demonstrated that both chronic and experimentally induced pain alter attentional performance in a task-dependent manner.

Pain assessment strategies span multiple modalities, each with distinct advantages and limitations. Self-report methods, such as numerical rating scales and standardized questionnaires, remain the clinical gold standard. They reflect the patient's own experience of pain, however, they when patients cannot communicate effectively, eg., critically ill, sedated, cognitively impaired, or pediatric populations they become less reliable. In these settings, clinicians usually rely on behavioral observation methods, such as facial expressions, vocalizations, and body movements, to estimate pain levels \cite{rojas_brown_2023}. In addition, physiological signals, including electrocardiography and skin conductance, have also been explored as more accurate indicators of nociceptive responses \cite{gkikas_tsiknakis_slr_2023}. However, these measures often capture broader stress or arousal responses rather than neural activity specific to pain. At the same time, developing effective analytical frameworks for the continuous monitoring and assessment of such data remains a wider challenge across domains \cite{chatziadam_dimitraidis_2020}.

Functional near-infrared spectroscopy (fNIRS) has emerged as a promising noninvasive neuroimaging modality for capturing cortical activity related to pain. It monitors changes in cerebral hemodynamics and oxygenation by simultaneously measuring concentrations of oxygenated hemoglobin (HbO) and deoxygenated hemoglobin (HbR) in the cortex \cite{rojas_huang_2016}.
Broadly, brain activity signals have been shown to reflect a range of affective and cognitive states \cite{gkikas_cruz_eeite_cwl_2026}.
Studies have consistently shown that noxious stimuli elicit distinct oxygenation changes across multiple cortical regions in both healthy subjects and clinical populations \cite{rojas_liao_2019}.

This work explores brain activity using fNIRS and evaluates two representations: the original 1D multichannel waveforms and their time-frequency spectrograms. We introduce a lightweight transformer-based architecture that performs multi-stream, multi-representation fusion through a specialized token-mixing mechanism, achieving stable performance while maintaining computational efficiency.

\section{Related Work}
Over the last decade, research on automatic pain assessment has grown rapidly. A transition from classical image processing pipelines to deep learning-based approaches has been made, although traditional methods still perform well, especially when data are limited \cite{gkikas_phd_thesis_2025}. Most existing methods focus on video, aiming to capture behavioral manifestations of pain through facial expressions, body motion, and other observable cues, and they employ a broad range of learning strategies \cite{huang_dong_2022,bargshady_hirachan_2024,gkikas_tsiknakis_embc,gkikas_tsiknakis_thermal_2024}. Although video-based approaches dominate the field, an increasing number of studies have also investigated biosignal-driven pain recognition, using physiological markers from electrocardiography \cite{gkikas_chatzaki_2022, gkikas_chatzaki_2023}, electromyography \cite{thiam_bellmann_kestler_2019,patil_patil_2024,pavlidou_tsiknakis_2025}, electrodermal activity (EDA) \cite{ji_zhao_li_2023,phan_iyortsuun_2023,lu_ozek_kamarthi_2023,li_luo_2024,aziz_joseph_2025,gkikas_kyprakis_eda_2025}, and brain activity measured with functional near-infrared spectroscopy \cite{rojas_huang_2016,rojas_liao_2019,rojas_romero_2021,khan_sousani_2024,rojas_joseph_bargshady_2024,bargshady_aziz_2025}.
Several pain studies have employed fNIRS in conjunction with machine learning to extract informative features and discriminate between pain conditions. In \cite{rojas_huang_2017_b}, a bag-of-words representation combined with a K-NN classifier on time--frequency features outperformed using time- or frequency-domain features alone. In contrast, \cite{rojas_juang_2019} reported the best results by combining time- and frequency-domain features with a Gaussian SVM, while \cite{rojas_romero_2021} used raw fNIRS signals with a two-layer BiLSTM, achieving $90.60\%$ accuracy in a multi-class classification task. Finally, in \cite{rojas_joseph_2024}, the authors proposed a hybrid CNN--LSTM architecture to capture spatio-temporal patterns in fNIRS and reported high performance.
Multimodal approaches to pain recognition have attracted increasing attention in the research community, since the combination of behavioral and physiological information can improve both robustness and predictive performance \cite{farmani_bargshady_2025, gkikas_tachos_2024, khan_chetty_2026}. In \cite{shi_chikhaoui_2022}, the authors fused features extracted from ECG and EDA and applied tree-based classifiers for pain recognition. Similarly, \cite{jerritta_murugappan_2022} combined wavelet-based features from ECG and EMG and used a k-nearest neighbor classifier. Beyond the use of biosignals \cite{liang_luo_2025}, a combination of facial and voice features in pediatric postoperative settings has shown strong performance.

Although the literature on automatic pain recognition continues to grow, most existing methods still process different modalities or signal representations through separate feature extraction or modeling pipelines. As a result, only a limited number of studies have explored architectural designs that handle heterogeneous inputs within a single unified framework. The present work addresses this by processing raw waveforms and time-frequency spectrograms through a single tokenization backbone, without representation-specific branches or specialized preprocessing.

\section{Methodology}
This section details the preprocessing pipeline for fNIRS signals and presents the proposed transformer model, along with its underlying tokenization approach.

\subsection{Pre-processing}
In the preprocessing stage, we use the fNIRS signals both in their original waveform form, represented as multichannel 1D sequences, and as spectrograms derived directly from the raw data. Every raw fNIRS channel is converted to a spectrogram by computing its power spectral density (PSD), which measures how the signal energy is distributed across frequencies over short overlapping windows. The result is a 2D time--frequency representation which is then resized to $224 \times 224$ pixels. Since the proposed framework does not rely on handcrafted feature extraction, it can be applied to either filtered or unfiltered signals. In this study, we use the unfiltered recordings, allowing the model to learn informative frequency patterns directly from the data while reducing the risk of removing pain-related spectral information through overly aggressive bandpass filtering.

\subsection{Architecture}
\label{sec:architecture}
The architecture uses the unified token representation produced by the proposed tokenization method, allowing a single model to process inputs with different dimensionalities, structures, or modalities.
A set of learnable latent vectors attends to the input tokens through cross-attention and is gradually refined through self-attention and feedforward layers \cite{gkikas_kyprakis_resp_2025}. This method allows the model to keep a fixed latent size while still handling differences in input resolution, sequence length, and modality.
The approach imposes a structured latent bottleneck by partitioning the token sequence into segments and associating a latent vector with each segment.
Specifically, for an input tensor:
\begin{equation}
\mathbf{X} \in \mathbb{R}^{A_1 \times \cdots \times A_D \times C},
\end{equation}
$D$ denotes the number of input axes, $A_1,\ldots,A_D$ the axis sizes, and $C$ the number of channels per input location. The number of axes $D$ is kept small (typically $D\in\{1,2\}$) by folding modality-specific factors into the channel dimension.
For 2D inputs, $D=2$ with $A_1 \times A_2 = H \times W$, and $\mathbf{X} \in \mathbb{R}^{H \times W \times C}$. If multiple image-like representations exist for a given sample --- fNIRS spectrograms being the case here --- we stack them along the channel axis, with $C$ counting the total channels.
For 1D inputs such as fNIRS waveforms, $D=1$ and $\mathbf{X} \in \mathbb{R}^{L \times C}$, where $L$ is the temporal length and $C$ the number of signal channels.

The input axes are flattened into a sequence of $N$ tokens, where $N$ equals the number of spatial or temporal positions (\textit{e.g.}, $N=H\times W$ for $D=2$ and $N=L$ for $D=1$). Each token is a vector comprising the input data features and positional embeddings. Positional information is encoded using Fourier features. For positions $\mathbf{p}\in[-1,1]^D$, the encoding with $K$ frequency bands and maximum frequency $f_{\max}$ is:
\begin{equation}
\gamma(\mathbf{p}) = \big[ \sin(\pi \mathbf{s} \mathbf{p}), \cos(\pi \mathbf{s} \mathbf{p}), \mathbf{p} \big],
\end{equation}
where $\mathbf{s} = [s_1, \ldots, s_K]^\top$ with $\{s_k\}_{k=1}^{K}$ spanning $[1, f_{\max}/2]$. After flattening, the data channels and positional features are concatenated to create the token matrix:
\begin{equation}
\mathbf{T} \in \mathbb{R}^{N \times C'}, \qquad
C' = C + D(2K+1).
\end{equation}
Multi-head attention (MHA) splits queries, keys, and values into $H$ heads, computes scaled dot-product attention per head, and concatenates the results:
\begin{equation}
\mathrm{MHA}(\mathbf{Q}, \mathbf{K}, \mathbf{V})
=
\mathrm{Concat}(\mathrm{head}_1,\ldots,\mathrm{head}_H)\mathbf{W}^O,
\end{equation}
where
\begin{equation}
\mathrm{head}_h
=
\mathrm{softmax}\!\left(
\frac{\mathbf{Q}_h \mathbf{K}_h^\top}{\sqrt{d_h}}
\right)\mathbf{V}_h,
\end{equation}
and $\mathbf{Q}_h=\mathbf{Q}\mathbf{W}^Q_h$, with analogous projections for keys and values.
Cross-attention and self-attention employ the same mechanism, but have different contexts:
\begin{equation}
\mathrm{MHA}(\mathbf{Q},\mathbf{K},\mathbf{V})
= \mathrm{MHA}(\mathbf{X}\mathbf{W}^Q,\ \mathbf{C}\mathbf{W}^K,\ \mathbf{C}\mathbf{W}^V),
\end{equation}
where queries come from $\mathbf{X}=\mathbf{L}$, and the context is $\mathbf{C}=\mathbf{T}$ for cross-attention or $\mathbf{C}=\mathbf{L}$ for self-attention.

\begin{figure*}
\begin{center}
\includegraphics[scale=0.68]{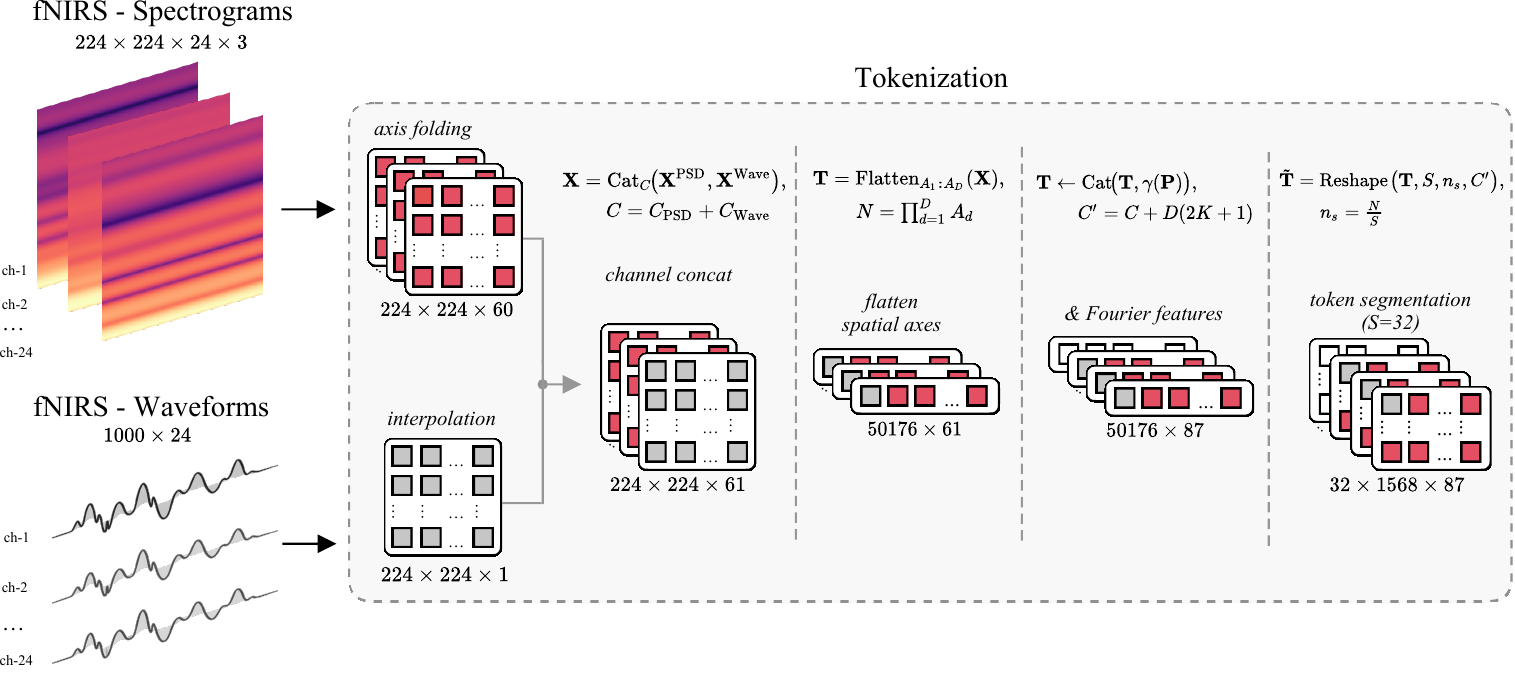} 
\end{center}
\caption{Overview of the proposed tokenization framework, using the two fNIRS representations.}
\label{overview}
\end{figure*}

The architecture maintains a structured set of latent states by partitioning the token sequence into $S$ contiguous segments, assigning one latent state to each segment. The number of latents $M$ reported in Table~\ref{tab:architecture_details} corresponds to $S$; it is fixed at $32$ for the fusion experiments but treated as a free parameter in the ablation study. Tokens are formed as described, yielding $\mathbf{T} \in \mathbb{R}^{N \times C'}$. The sequence is segmented into $S$ segments of length $n_s=\lceil N/S \rceil$, padded to $\tilde{N}=Sn_s$, and reshaped as:
\begin{equation}
\tilde{\mathbf{T}} \in \mathbb{R}^{S \times n_s \times C'}.
\end{equation}
The tokens of segment $s$ are denoted $\tilde{\mathbf{T}}_s \in \mathbb{R}^{n_s \times C'}$.
At runtime, all segment states start from the same learnable vector $\boldsymbol{\ell}_{\mathrm{init}} \in \mathbb{R}^{d}$ --- there are no separate parameters per segment. What differentiates them is purely the result of local attention: each state is updated by attending over its own assigned tokens, and segment-specific structure builds up layer by layer. Each layer applies a cross-attention and is followed by a feedforward layer, with pre-normalization and residual connections. The update for segment $s$ at layer $\ell$ is defined as:
\begin{equation}
\mathbf{e}^{(\ell)}_s
=
\mathbf{e}^{(\ell-1)}_s
+
\mathrm{Attn}\!\big(\mathbf{e}^{(\ell-1)}_s,\ \tilde{\mathbf{T}}_s\big),
\end{equation}
where $\mathbf{e}^{(\ell)}_s \in \mathbb{R}^{d}$ and $\mathrm{Attn}(\mathbf{L},\mathbf{C})$ denotes a pre-normalized multi-head attention block with queries from $\mathbf{L}$ and key-values from $\mathbf{C}$, followed by a feedforward layer. By stacking segment states results to $\mathbf{E}^{(\ell)} \in \mathbb{R}^{S \times d}$. Inter-segment communication is happening by applying self-attention on the segment states:
\begin{equation}
\mathbf{E}^{(\ell)} \leftarrow
\mathbf{E}^{(\ell)} +
\mathrm{Attn}\ \!\big(\mathbf{E}^{(\ell)},\ \mathbf{E}^{(\ell)}\big),
\end{equation}
applied multiple times per layer to enhance the interactions between latents. Local cross-attention is parallelized by treating segments as batch elements, without changing the computation. After $L$ layers, the final segment, $\mathbf{E}^{(L)}$, provides the representation for prediction.
Figure \ref{overview} illustrates the proposed architecture in a multimodal setup, fusing fNIRS waveforms and their spectrograms.
Model hyperparameters for all experiments are summarized in Table \ref{tab:architecture_details}, while Table \ref{table:training_details} presents the training setup.

%%%%%%%%%%%%%%%%%%%%%%%%%%%%%%%%%%%%%%%%%%%%%%%%%%%%%%%%%%%%%%%%%%%%%%%%%%%%%%%%%%%%%%%

\begin{table}[t]
\centering
\footnotesize
\caption{Architectural details.}
\label{tab:architecture_details}
\begin{tabular}{l c}
\toprule
\textbf{Hyperparameter} & \textbf{Value} \\
\midrule
Depth ($L$) & 4 \\
Number of latents ($M$) & 32 \\
Latent dimension ($d$) & 128 \\
Cross-attention heads & 1 \\
Latent self-attention heads & 8 \\
Cross-attention head dimension & 64 \\
Latent attention head dimension & 64 \\
Self-attention blocks per cross ($R$) & 8 \\
\bottomrule
\end{tabular}
\end{table}

%%%%%%%%%%%%%%%%%%%%%%%%%%%%%%%%%%%%%%%%%%%%%%%%%%%%%%%%%%%%%%%%%%%%%%%%%%%%%%%%%

\begin{table}
\scriptsize
\caption{Training configuration.}
\label{table:training_details}
\begin{center}
\begin{threeparttable}
\begin{tabular}{ P{0.75cm} P{0.45cm}  P{0.60cm}  P{0.65cm} P{0.65cm} P{0.65cm} P{0.80cm} P{0.40cm}}
\toprule
\multirow{2}{*}{\shortstack{Optimizer}} &
\multirow{2}{*}{\shortstack{LR}} &
LR decay &Weight decay &Total epochs &Warmup epochs &Cooldown epochs &Batch size\\
\midrule
\midrule
\textit{AdamW} &\textit{2e-5} &\textit{cosine} &0.1 &200 &20 &10 &32 \\
\bottomrule
\end{tabular}
\end{threeparttable}
\end{center}
\end{table}

\section{Experimental Evaluation \& Results}
\label{sec:results}
This study uses the AI4Pain dataset \cite{ai4pain_2024,rojas_hirachan_2023}, which comprises facial videos and fNIRS recordings from $65$ participants. The data were recorded at the Human--Machine Interface Laboratory at the University of Canberra, Australia, and divided into $41$ subjects for training, $12$ for validation, and $12$ for testing. Pain was induced using transcutaneous electrical nerve stimulation, with electrodes placed on the inner forearm and the back of the right hand.
Pain threshold denotes the lowest stimulus intensity perceived as painful (low pain), whereas pain tolerance denotes the highest intensity a participant can tolerate before it becomes unbearable (high pain).
For the fNIRS modality, $24$ oxygenated hemoglobin (HbO) channels were employed. All experiments use a three-class classification scheme (No Pain, Low Pain, High Pain) and are evaluated under a hold-out strategy on the validation set, reporting macro-averaged accuracy, precision, and F1 score. Comparisons with previous methods on the testing set are detailed in Section \ref{comparison_testing}.

\subsection{Impact of Latent Segmentation on fNIRS Representations}

Table \ref{table:v3_2_segments} reports performance and computational characteristics of the proposed framework on fNIRS inputs across different latent segmentation settings, considering both raw waveform and spectrogram-based PSD representations using all the available $24$ channels. Changing the number of segments affects how tokens are grouped into latent states, without changing their dimensionality. This mainly influences the computational cost and the locality of the learned representations.

For the waveform representation, accuracy is largely stable across segment settings, ranging from $43.94\%$ without segmentation to $47.18\%$ at $32$ segments. The low performance of this representation shows that the raw waveform provides limited discriminative information, regardless of the latent organization. From a computational perspective, the waveforms remain a lightweight modality.  Specifically, GFLOPs increase gradually with the number of segments, whereas GPU latency stays nearly unchanged at about $13$--$14$ ms across all settings.

Across all segmentation settings, the PSD representation performs better than the raw waveform, with accuracy above $49\%$ and reaches a peak at $52.31\%$ for $S=2$. This indicates that representing the signal in the time--frequency domain provides more informative pain-related hemodynamic information than using the raw waveform. However, the difference may also partly reflect the higher-dimensional tokenization of PSD inputs. Although PSD incurs higher GFLOPs, GPU latency remains comparable to waveform processing --- a practically important property for deployment.

\begin{figure}
\begin{center}
\includegraphics[scale=0.56]{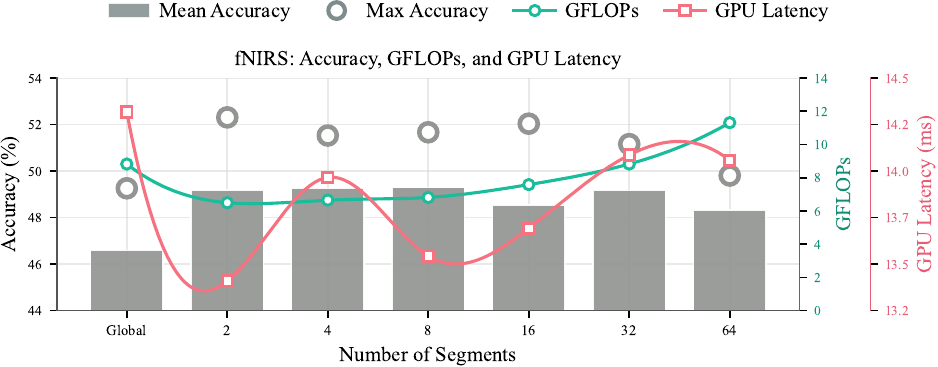}
\end{center}
\caption{Effect of latent segmentation on fNIRS accuracy (averaged across waveform and spectrogram representations) versus computational and inference cost.}
\label{accuracy_cost}
\end{figure}

Figure \ref{accuracy_cost} presents the mean accuracy, which is the average across waveform and spectrogram representations, alongside GFLOPs and GPU latency for the latent segmentation settings reported in Table \ref{table:v3_2_segments}. We observe that as the number of segments increases, the computational cost rises gradually, while recognition accuracy remains largely stable, with limited sensitivity to the segmentation level. Segment numbers between $2$ and $16$ yield lower GFLOPs and reduced GPU latency than the unsegmented global representation, while maintaining comparable performance.

\subsection{Representation Fusion}
\label{sec:modality_fusion}

The effect of combining the two fNIRS representations is evaluated by stacking the waveform and PSD tensors along the channel dimension. This enables a joint modeling with a single shared backbone. 
For this analysis, $S=32$ segments are selected, as this configuration provides strong unimodal performance with balanced computational and inference costs. The fusion results are reported in Table \ref{table:fusion}.

Using the stacking approach for the waveform and PSD representations led to $48.52\%$ accuracy, $45.62\%$ precision, and $44.62\%$ F1 score. Although this results in performance better than the highest performing PSD-only model ($52.31\%$ with $S = 2$), we demonstrate that both representations may be used together as part of a common backbone for modeling without changing the underlying architecture. We further show that by combining representations using our method, the raw waveform does not provide additional information about the signal beyond what is contained in its PSD. 

We observe that the increase in computational cost resulting from adding the waveform representation remains small compared to PSD-only processing, and similarly, inference latency has remained very similar to the PSD-only case on both the GPU and CPU.
The representation fusion is perfomed by using the 1D waveform grid which is projected onto a 2D layout through interpolation and concatenated with the PSD tensor along the channel dimension:
\begin{equation}
\mathbf{X}^{\text{stack}}=\mathrm{Cat}_{C}\big(\mathcal{I}(\mathbf{X}^{\text{wave}}),\ \mathbf{X}^{\text{PSD}}\big),
\end{equation}
where $\mathcal{I}$ denotes interpolation from the $C\times L$ waveform structure to a single $H\times W$ tensor.

%%%%%%%%%%%%%%%%%%%%%%%%%%%%%%%%%%%%%%%%%%%%%%%%%
\begin{table*}
\scriptsize
\caption{performance and computational/inference cost of fNIRS representations across latent segmentation settings.}
\label{table:v3_2_segments}
\begin{center}
\begin{threeparttable}
\begin{tabular}{P{1.3cm} P{0.60cm} P{1.1cm} P{1.1cm} P{2.1cm} P{1.8cm} P{2.1cm} P{1.7cm} P{0.7cm} P{0.6cm} P{0.4cm}}
\toprule

\multirow{2}{*}{\shortstack{Input}} &
\multirow{2}{*}{\shortstack{\#Segm.}} &
\multicolumn{2}{c}{Computational Cost} &
\multicolumn{4}{c}{Inference Cost} &
\multicolumn{3}{c}{Performance} \\

\cmidrule(lr){3-4}\cmidrule(lr){5-8}\cmidrule(lr){9-11}
& & Params(M) & GFLOPs &Latency (ms) GPU$\downarrow$  &Samples/s GPU$\uparrow$ &Latency (ms) CPU$\downarrow$  &Samples/s CPU$\uparrow$  & Accuracy & Precision & F1 \\

\midrule
\midrule

fNIRS-Wave &-- &7.81  &2.58  &14.26 &70.14 &22.44  &44.56 &43.94 &42.94 &39.34 \\
fNIRS-PSD  &-- &15.05 &15.07 &14.36 &69.64 &56.14  &17.81 &49.26 &\textbf{47.91} &39.90\\\hdashline

fNIRS-Wave &2 &7.81  &0.25  &13.33 &74.99 &11.33  &88.27 &46.06 &42.63 &43.20\\
fNIRS-PSD  &2 &15.05 &12.74 &13.40 &74.62 &51.03  &19.59 &\textbf{52.31} &\underline{46.66} &\textbf{47.76}\\\hdashline

fNIRS-Wave &4 &7.81  &0.41  &14.31 &69.90 &11.80  &84.75 &46.99 &44.32 &43.59\\
fNIRS-PSD  &4 &15.05 &12.90 &13.58 &73.63 &45.02  &22.21 &51.53 &46.12 &\underline{47.11}\\\hdashline

fNIRS-Wave &8 &7.81  &0.41  &13.62 &73.39 &13.80  &72.48 &46.90 &43.74 &42.96\\
fNIRS-PSD  &8 &15.05 &13.21 &13.39 &74.70 &45.98  &21.75 &51.67 &49.58 &43.06 \\\hdashline

fNIRS-Wave &16 &7.81  &1.34  &13.48 &74.16 &17.95  &55.72 &45.05 &42.64 &39.42\\
fNIRS-PSD  &16 &15.05 &13.83 &13.84 &72.25 &46.11  &21.69 &\underline{52.04} &46.59 &45.88\\\hdashline

fNIRS-Wave &32 &7.81  &2.58  &13.50 &74.06 &22.97  &43.54 &47.18 &43.54 &41.34\\
fNIRS-PSD  &32 &15.05 &15.07 &14.64 &68.31 &48.88  &20.46 &51.16 &45.42 &45.70\\\hdashline

fNIRS-Wave &64 &7.81  &5.07  &14.09 &70.99 &31.94  &31.31 &46.85 &42.19 &41.81\\
fNIRS-PSD  &64 &15.05 &17.56 &13.99 &71.48 &51.83  &19.29 &49.81 &29.01 &36.67 \\

\bottomrule
\end{tabular}

\begin{tablenotes}[para,flushleft]
\scriptsize
\emph{Notes:} For fNIRS, we use all available channels ($24$). The \textbf{bold} font marks the highest score within each modality, while the \underline{underline} marks the second-highest. \space Segm.: Segments \space 
\end{tablenotes}
\end{threeparttable}
\end{center}
\end{table*}
%%%%%%%%%%%%%%%%%%%%%%%%%%%%%%%%%%%%%%%%%%%%%%%%%

%%%%%%%%%%%%%%%%%%%%%%%%%%%%%%%%%%%%%%%%%%%%%%%%%

%%%%%%%%%%%%%%%%%%%%%%%%%%%%%%%%%%%%%%%%%%%%%%%%

\begin{table*}
\scriptsize
\caption{performance and computational/inference cost for the representation fusion.}
\label{table:fusion}
\begin{center}
\begin{threeparttable}
\begin{tabular}{P{1.3cm} P{0.60cm} P{0.95cm} P{0.95cm} P{2.1cm} P{1.8cm} P{2.1cm} P{1.7cm} P{0.70cm} P{0.60cm} P{0.35cm}}
\toprule

\multirow{2}{*}{\shortstack{Modality}} &
\multirow{2}{*}{\shortstack{Fusion}} &
\multicolumn{2}{c}{Computational Cost} &
\multicolumn{4}{c}{Inference Cost} &
\multicolumn{3}{c}{Performance} \\

\cmidrule(lr){3-4}\cmidrule(lr){5-8}\cmidrule(lr){9-11}
 & & Params(M) & GFLOPs &Latency (ms) GPU$\downarrow$  &Samples/s GPU$\uparrow$ &Latency (ms) CPU$\downarrow$  &Samples/s CPU$\uparrow$  & Accuracy & Precision & F1 \\

\midrule
\midrule

fNIRS-Wave &\multirow{2}{*}{\centering stack}  &\multirow{2}{*}{\centering 15.15}  &\multirow{2}{*}{\centering 15.20}  &\multirow{2}{*}{\centering 13.79}  &\multirow{2}{*}{\centering 72.54}  &\multirow{2}{*}{\centering 49.00}  &\multirow{2}{*}{\centering 20.41}  &\multirow{2}{*}{\centering 48.52}  &\multirow{2}{*}{\centering 45.62} &\multirow{2}{*}{\centering 44.62} \\
fNIRS-PSD  &  &  &  &  &  &  &  &  &  &  \\

\bottomrule
\end{tabular}
\end{threeparttable}
\end{center}
\end{table*}

%%%%%%%%%%%%%%%%%%%%%%%%%%%%%%%%%%%%%%%%%%%%%%%%

%%%%%%%%%%%%%%%%%%%%%%%%%%%%%%%%%%%%%%%%%%%%%%%%%%%%%%%%%%%%%%%%%%%%%%%%%%%%%%%
\section{Comparison with Existing Methods}
\label{comparison_testing}

In Table \ref{table:ai4pain_test} we compare the performance of all three approaches using data from the \textit{AI4Pain} test set: Video-based, fNIRS-based, and multimodal approaches. Handcrafted feature-based traditional methods were shown to have limited capacity for representing complex patterns associated with pain; even so, the best accuracy reported by an SVM approach presented in \cite{ai4pain_2024} was $43.30\%$, regardless of modality.
From the othe hand deep learning methods generally perform better. The multimodal transformer in \cite{gkikas_tsiknakis_painvit_2024} achieves $46.67\%$, the video-only CNN in \cite{prajod_schiller_2024} reaches $49.00\%$, while the CNN--Transformer fusion model in \cite{vianto_2025} $51.33\%$. Using handcrafted fNIRS features, the ensemble classifier in \cite{khan_aziz_2025} the authors attains $53.66\%$, while the video transformer in \cite{nguyen_yang_2024} reports $55.00\%$. The strongest performance is presented by a multimodal transformer in \cite{gkikas_rojas_painformer_2025}, which achieves $55.69\%$ combining facial video and fNIRS.

The proposed method reaches $50.00\%$ accuracy on the test set with the Wave+PSD stack fusion. Even though it relies only on brain activity signals, it outperforms the SVM-based fNIRS baselines. It also remains competitive with several deep learning methods that include behavioral information, which supports the potential of unified tokenization and representation fusion for fNIRS-based pain recognition.

%%%%%%%%%%%%%%%%%%%%%%%%%%%%%%%%%%%%%%%%%%%%%%%%
\begin{table}
\footnotesize
\caption{Comparison of studies on the \textit{AI4Pain} testing set.}
\label{table:ai4pain_test}
\begin{center}
\begin{threeparttable}
\begin{tabular}{P{0.6cm} P{0.7cm} P{0.7cm} P{1.2cm} P{2.1cm} P{0.9cm}}
\toprule
\multirow{2}{*}{\shortstack{Study}} &
\multicolumn{2}{c}{Modality} &
\multicolumn{2}{c}{Method} &
\multirow{2}{*}{\shortstack{Accuracy}} \\
\cmidrule(lr){2-3}\cmidrule(lr){4-5}
& Video & fNIRS &Features &Model & \\
\midrule
\midrule

& ---        & \checkmark &   &     &43.30\\
& \checkmark & ---        &   &     &40.10\\
\multirow{-3}{*}{\cite{ai4pain_2024}}
& \checkmark & \checkmark &\multirow{-3}{*}{Handcrafted} & 
\multirow{-3}{*}{SVM} &41.70\\\hdashline
%%%%%%%%%%%%%%%%%%%%%%%%%%%%%%%%%%%%%%%%%%%%%%%%%%%%%%%
\cite{gkikas_tsiknakis_painvit_2024} & \checkmark & \checkmark  &Deep &Transformer      &46.67\\ \hdashline
\cite{prajod_schiller_2024}          & \checkmark & ---         &Deep & 2D CNN          &49.00\\ \hdashline
\cite{vianto_2025}                   & \checkmark & \checkmark  &Deep & CNN-Transformer &51.33\\ \hdashline
\cite{khan_aziz_2025}                & ---        & \checkmark  &Handcrafted & ENS      &53.66\\ \hdashline
\cite{nguyen_yang_2024}              & \checkmark & ---         &Deep & Transformer     &55.00\\ \hdashline

%%%%%%%%%%%%%%%%%%%%%%%%%%%%%%%%%%%%%%%%%%%%%%%%%%%%%%%
& ---        & \checkmark &    &    & 52.60\\
& \checkmark & ---        &    &    & 53.67\\
\multirow{-3}{*}{\cite{gkikas_rojas_painformer_2025}}
& \checkmark & \checkmark &\multirow{-3}{*}{Deep} & 
\multirow{-3}{*}{Transformer} &55.69\\\midrule
%%%%%%%%%%%%%%%%%%%%%%%%%%%%%%%%%%%%%%%%%%%%%%%%%%%%%%%

\rowcolor{mygray}
Our & ---        & \checkmark &   Deep  & Transformer  & 50.00$^\dagger$ \\

\bottomrule
\end{tabular}
\begin{tablenotes}
\scriptsize
\item \checkmark indicates the modality is used \space -- indicates not used \space ENS: Ensemble Classifier 
\item $\dagger$: Wave \& PSD stack fusion
\end{tablenotes}
\end{threeparttable}
\end{center}
\end{table}
%%%%%%%%%%%%%%%%%%%%%%%%%%%%%%%%%%%%%%%%%%%%%%%%

\section{Conclusion}
In this study, we present a lightweight transformer for fNIRS-based pain recognition that combines unified tokenization with latent segmentation, allowing multiple signal representations to be processed by a single model. Experiments on the \textit{AI4Pain} dataset showed that the proposed model maintained a stable performance across a range of computational settings.
Fusing raw waveform and PSD representations demonstrated the feasibility of joint processing within a single backbone. 
These findings support the viability of efficient transformer designs for neurophysiological pain assessment in constrained settings~\cite{gkikas_tiny_2025, antonogiorgakis_britzolakis_2019}, including clinical and interactive environments~\cite{10807220} where social robots could use real-time estimates of pain and cognitive states to adapt their behavior~\cite{arzate_when_2025, arzate_hri_2025, arzate_3dmm_2022, fang_moleron_2024, kruger_oshima_2026, hessels_fang_2026,vazquez_cruz_gkikas_2026}.

\bibliographystyle{IEEEtran}
\bibliography{library}

\end{document}